\documentclass[letterpaper, 10 pt, conference]{ieeeconf} 
\usepackage{amsmath,amssymb,amsfonts}
\usepackage{graphicx}
\usepackage{algpseudocode}
\usepackage[linesnumbered,ruled,vlined]{algorithm2e}
\usepackage{pifont}
\usepackage{hyperref}
\usepackage{comment}
\usepackage{todonotes}
\usepackage{subcaption}
\usepackage{float}
\usepackage{placeins}
\usepackage{multirow}

\IEEEoverridecommandlockouts

\title{\LARGE \bf PRISM-Loc: a Lightweight Long-range LiDAR Localization in Urban Environments with Topological Maps}

\author{Kirill Muravyev$^{1,2}$, Artem Kobozev$^{2}$, Vasily Yuryev$^{3}$, Alexander Melekhin$^{3}$, Oleg Bulichev$^{3,5}$,\\Dmitry Yudin$^{3,4}$, and Konstantin Yakovlev$^{1,4}$
\thanks{$^{1}$Kirill Muravyev and Konstantin Yakovlev are with Federal Research Center for Computer Science and Control of Russian Academy of Sciences,
        {\tt\small \{muraviev,yakovlev\}@isa.ru}}%
\thanks{$^{2}$Kirill Muravyev and Artem Kobozev are with National Research University "Higher School of Economics",
        {\tt\small kmuravev@hse.ru, avkobozev@edu.hse.ru}}%
\thanks{$^{3}$Vasily Yuryev, Alexander Melekhin, Oleg Bulichev, and Dmitry Yudin are with Moscow Institute of Physics and Technology (MIPT)
        {\tt\small \{yuryev.ve,melekhin.aa\}@phystech.edu, bulichev.ov@mipt.ru}}%
\thanks{$^{4}$ Dmitry Yudin and Konstantin Yakovlev are also with Artificial Intelligence Research Institute (AIRI)
        {\tt\small \{yudin.da,yakovlev\}@airi.net}}%
\thanks{$^{5}$ Oleg Bulichev is also with Innopolis University
        {\tt\small {o.bulichev}@innopolis.ru}}%
}

\begin{document}

\maketitle
\thispagestyle{empty}
\pagestyle{empty}



\begin{abstract}
We propose PRISM-Loc -- a lightweight and robust approach for localization in large outdoor environments that combines a compact topological representation with a novel scan-matching and curb-detection module operating on raw LiDAR scans. The method is designed for resource-constrained platforms and emphasizes real-time performance and resilience to common urban sensing challenges. It provides accurate localization in compact topological maps using global place recognition and an original scan matching technique. Experiments on standard benchmarks and on an embedded platform demonstrate the effectiveness of our approach. Our method achieves a 99\% success rate on the large-scale ITLP-Campus dataset while running at 150 ms per localization and using a 20 MB map for localization. We highlight three main contributions: (1) a compact representation for city-scale localization; (2) a novel curb detection and scan matching pipeline operating directly on raw LiDAR points; (3) a thorough evaluation of our method with performance analysis.
\end{abstract}

\begin{figure}
    \centering
    \includegraphics[width=0.45\textwidth]{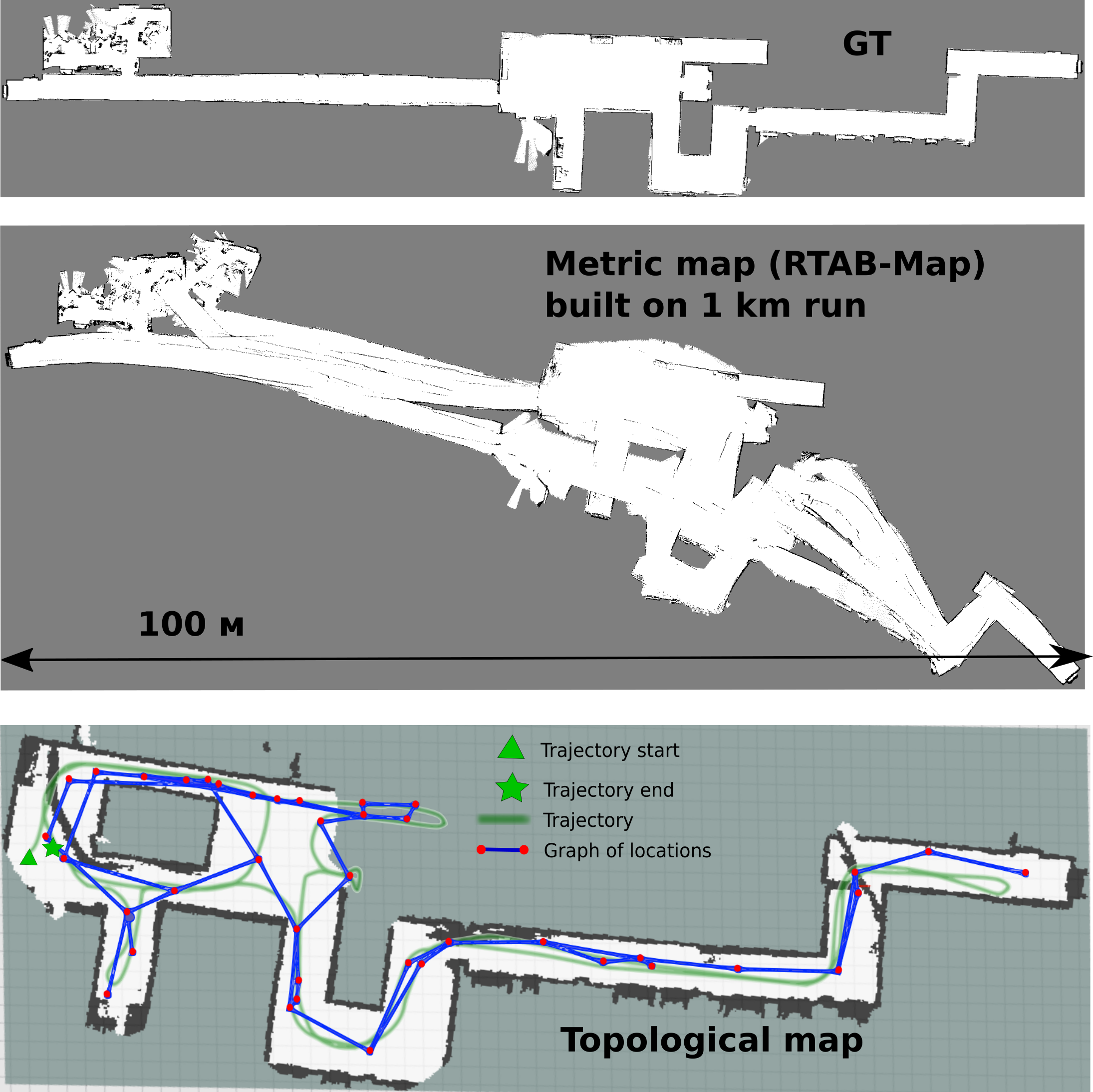}
    \caption{Ground truth metric map (top), metric map built by RTAB-Map (middle), and topological map aligned with a ground-truth metric map (bottom).}
    \label{fig:corridor_bifurcation}
\end{figure}

\section{INTRODUCTION}
Localization is a fundamental requirement for mobile robots and self-driving vehicles, enabling them to accurately determine their position within complex environments. Traditionally, high-precision localization has relied on dense global LiDAR maps that provide detailed geometric representations of the surroundings. However, as the operational range increases, these dense maps become computationally expensive to process in real-time and require substantial memory resources. This challenge motivates the exploration of alternative representations, such as topological maps, which capture essential structural information with significantly lower memory demands.

Topological simultaneous localization and mapping (SLAM) methods represent the environment as a graph-based structure, where nodes correspond to key locations and edges define navigable paths between them. Unlike metric SLAM, which focuses on precise spatial coordinates, topological SLAM emphasizes structural and relational aspects of the environment. This approach enables more scalable localization while reducing computational and storage requirements.

Despite utilizing a graph-based structure, our approach still requires solving a rough pose estimation problem. This is achieved by matching incoming LiDAR scans against a database of precomputed topological nodes, each represented by a lightweight descriptor. This process yields a candidate location that approximates the vehicle’s global position.

In the second stage, we refine this estimation using a novel LiDAR scan matching algorithm. This algorithm, based on 2D feature extraction and point-based optimization, accurately determines the local pose within the identified candidate region. To increase accuracy of matching in urban environments, an original algorithm of curbs detection is used.

In this study, we compare the proposed method against two state-of-the-art localization techniques using three large-scale outdoor urban datasets. The first method relies solely on point cloud data for localization, while the second integrates place recognition and scan matching to enhance accuracy and robustness. To evaluate the efficiency of these approaches, we analyze their computational and memory performance, comparing them with our proposed method. This assessment provides insights into the trade-offs between accuracy, efficiency, and scalability in large-scale urban environments, highlighting the advantages of our approach for real-world robotic applications.

To sum up, the contributions of this work are as follows:
\begin{itemize}
    \item Precise and lightweight localization method in topological maps which is based on place recognition, scan matching, and rough odometry estimation.
    \item A novel LiDAR scan matching algorithm which estimates a precise relative transformation between two scans using the curbs detection.
    \item An original algorithm of curbs detection based on raw LiDAR point clouds.
\end{itemize}

\section{RELATED WORK}
\label{sec:related_work}

\subsection{Long-range Mapping}

The construction of large-scale metric maps for mobile robots has been extensively studied. One of the most common mapping methods is RTAB-Map~\cite{labbe2019rtab}, which employs a unique memory management technique for large-scale mapping. However, an evaluation conducted in~\cite{muravyev2022evaluation} shows that, over long distances and with a noisy positioning source, RTAB-Map’s memory consumption increases significantly (up to 8 GB for a 1 km trajectory) and produces inconsistent maps with artifacts such as "corridor bifurcation" (see Fig.~\ref{fig:corridor_bifurcation}).

To mitigate memory and computational costs, metric submap-based methods such as Voxgraph~\cite{reijgwart2019voxgraph} can be utilized. The GLIM method~\cite{koide2024glim} constructs a highly accurate metric map by representing the environment as a set of submaps, which are then stitched together into a voxelized point cloud. The authors tested this method on a 2 km trajectory in a park, achieving good trajectory estimation quality in real-time. However, memory consumption was not addressed in their study.

An alternative approach for long-range, large-scale mapping is the use of topological maps instead of dense global metric maps. This strategy has been implemented in works such as~\cite{niijima2020city} and~\cite{gomez2020hybrid}. In~\cite{gomez2020hybrid}, a topological map is represented as a graph of rooms, where doors serve as edges, and a metric map is constructed for each room separately. In~\cite{niijima2020city}, a city-scale topological map is built by fusing 2D global open-source maps with 3D point clouds obtained by a robot. This method can map a 17 $km^2$ urban indoor-outdoor environment while consuming only 1 GB of RAM. However, neither work provides a detailed description of the localization procedure within the topological map.

The TLF method~\cite{tang2019topological} constructs a topological map based on multiple runs, facilitating lifelong localization under varying daytime, weather, and dynamic conditions. Localization is performed using a stereo camera, achieving an accuracy of 0.5 m and 4 degrees over multiple runs in different environments.

In recent years, learning-based topological mapping methods have emerged. For example,~\cite{wiyatno2022lifelong} introduces a framework for lifelong navigation, localization in a pre-built topological map, and map updates to accommodate dynamic changes. Localization is performed using a neural network that compares the current image with reference images from the topological graph. While such methods achieve high navigation success rates using stereo or RGB-D cameras, they are typically restricted to indoor environments and short trajectories.

\subsection{Place Recognition}

To localize within a topological map, the system must first identify the correct location. This task is typically performed using place recognition techniques, which search for the location in the map whose observation best matches the current observation.

Early place recognition methods relied on the bag-of-words approach, where each observation is represented as an unordered set of descriptors. In visual place recognition, one of the most common bag-of-words-based methods is DBoW2~\cite{galvez2012bags}. In the context of point cloud-based place recognition, BoW3D~\cite{cui2022bow3d} has been successfully applied for loop closure on large outdoor trajectories.

More recently, learning-based place recognition methods have gained popularity. These methods encode an input image or point cloud as a single feature vector (descriptor) using a neural network. The correct location is then identified by comparing descriptor similarity between the current observation and map entries.

For vision-based place recognition, methods such as NetVLAD~\cite{arandjelovic2016NetVLADCNNArchitecturea}, CosPlace~\cite{berton2022RethinkingVisualGeolocalizationa}, and MixVPR~\cite{ali-bey2023MixVPRFeatureMixinga} use convolutional or transformer-based architectures to predict descriptors. These methods achieve a Top-1 place recognition recall above 0.85 in outdoor urban environments.

For point cloud-based place recognition, PointNetVLAD~\cite{uy2018PointNetVLADDeepPoint} extends the NetVLAD architecture to 3D data. Other methods, such as MinkLoc3D~\cite{komorowski2021MinkLoc3DPointCloud} and its enhancement~\cite{komorowski2022ImprovingPointCloudb}, utilize voxelization and sparse convolutions. SVT-Net~\cite{fan2022SVTNetSuperLightWeight} further refines this process by incorporating attention mechanisms. These approaches achieve recall rates of 0.87–0.93 in outdoor place recognition for urban environments.

Multimodal place recognition methods have also been developed, integrating image and point cloud data to improve performance. Examples include MinkLoc++~\cite{komorowski2021MinkLocLidarMonoculara} and MSSPlace~\cite{mssplace}. In our work, we utilize MinkLoc3D for place recognition as a lightweight and real-time method that relies solely on LiDAR point clouds.

\subsection{Pose Estimation}

Once the correct location is identified through place recognition, pose estimation is performed using image or scan matching techniques. These methods extract features from the robot’s current observation and place recognition results, estimating the transformation between them.

For vision-based pose estimation, an example is StereoScan~\cite{geiger2011stereoscan}, which determines the transformation using stereo image pairs.

For point cloud-based pose estimation, a common approach is Iterative Closest Point (ICP)~\cite{besl1992method}, which refines the transformation between two point clouds given an initial rough estimate. The RANSAC algorithm~\cite{fischler1981random} is often used to generate this initial guess. However, the RANSAC + ICP pipeline is computationally expensive for dense point clouds and performs well only when the input point clouds have significant overlap. According to~\cite{muravyev2025prism}, this pipeline achieves a recall of 0.75 for point cloud pairs with overlap above 50\%, but drops to 0.53 for pairs with overlap above 25\%.

Using 2D projections of point clouds (e.g., bird’s-eye views) can improve pose estimation accuracy when overlap is low while also accelerating the matching process compared to full 3D point clouds. Methods such as BVMatch~\cite{luo2021bvmatch} and its enhancement BEVPlace++~\cite{luo_BEVPlace_2024} introduce a density-based bird’s-eye-view extraction algorithm and a novel local descriptor, achieving a pose estimation error of 0.4 m.

In our previous work~\cite{muravyev2025prism}, we introduced a fast bird’s-eye-view matching algorithm based on ORB feature matching with iterative outlier removal. In this work, we enhance our scan matching algorithm for outdoor environments by incorporating curb detection in point clouds and point-to-distance-map optimization.

\section{PROBLEM STATEMENT}
\label{sec:problem_statement}

Consider a mobile robot navigating in an outdoor urban environment using a pre-built map. During navigation, the robot does not have access to a global positioning source such as GNSS. Instead, it receives odometry information and point clouds from its LiDAR sensors at each time step. The objective is to estimate the robot's position in the environment at each moment based on its sensor observations and the pre-built map. The robot's initial position in the map is assumed to be known.

Formally, the localization problem is defined as follows:

During the first run, we collect input point clouds $C_t^{\text{first}}$ and the corresponding ground truth robot positions $p_t^{\text{first}}$ for each time step $t = 1, \dots, T^{F}$. In subsequent runs, the robot receives new point clouds $C_t^{\text{next}}$ and odometry measurements $o_t^{\text{next}}$, which are subject to noise, for each time step $t = 1, \dots, T^{N}$. The initial position $p_0^{\text{next}}$ is known, while the ground truth positions $p_t^{\text{next}}$ for $t = 1, \dots, T^N$ remain unknown.

After the first run, we construct a map $M$ using the collected data:
$$M = Mapping(C_1^{first}, \dots, C_{T^{F}}^{first}; p_1^{first}, \dots, p_{T^{F}}^{first})$$

The localization task at each step $t$ is as follows:

$$\widehat{p_{t}^{next}} = Loc(M; C_t^{next}; o_t^{next}; \widehat{p_{t-1}^{next}})$$.

As the localization quality metric, we use the Absolte Trajectory Error (ATE). We compute mean and median error value over all the time steps:

\begin{equation}
e_t = ||\widehat{p_{t}^{next}} - p_{t}^{next}||_2,
\end{equation}

\begin{equation}
ATE_{mean} = \frac{1}{T^N} \sum\limits_{t=1}^{T^N} e_t,
\label{eq:ate_mean}
\end{equation}

\begin{equation}
ATE_{median} = Median(e_1, \dots, e_{T^N}).
\label{eq:ate_median}
\end{equation}

Also we evaluate the localization success rate as the part of time steps where the localization error is lower than a certain threshold:
\begin{equation}
SR_{loc} = \frac{1}{T^N} \sum\limits_{t=1}^{T^N} I(e_t < Th).
\label{eq:sr_loc}
\end{equation}

\FloatBarrier

\section{METHOD}

\subsection{Notation}
We summarize commonly used notation. Scalars are denoted by lowercase
letters (e.g., $t$), vectors by bold lowercase (e.g., $\mathbf{x}$), and matrices by uppercase letters
(e.g., $T$). If you already defined notation later, please move the definitive description here.

\label{sec:method}

\subsection{Map creation}

\begin{figure}
    \vspace*{0.15in}
    \centering
    \includegraphics[width=0.45\textwidth]{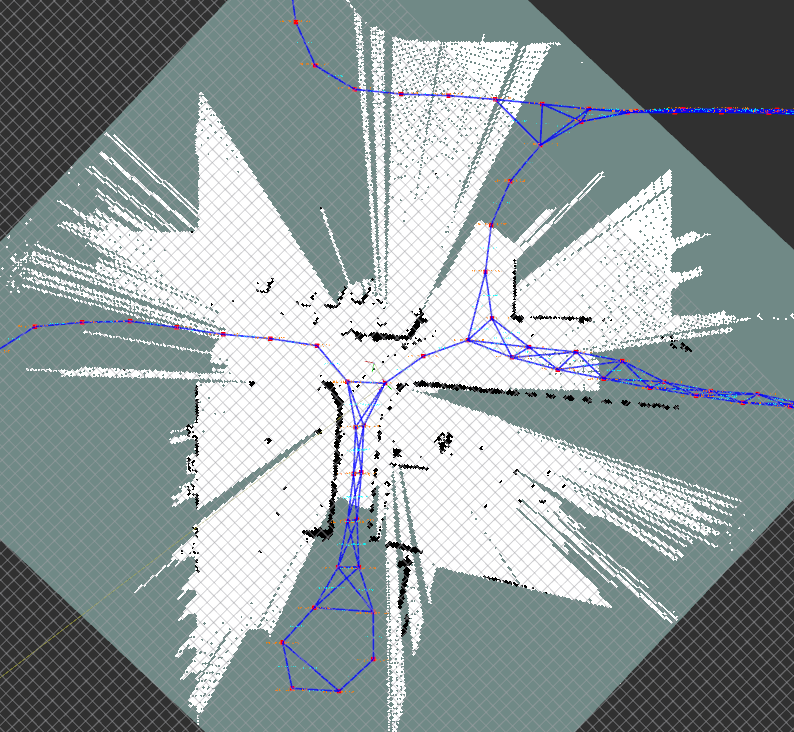}
    \caption{An example of a topological map built by PRISM-TopoMap method (red nodes and blue edges), and a 2D grid of a location.}
    \label{fig:topomap}
\end{figure}

For map creation, we use the concept of a graph of locations and its implementation from PRISM-TopoMap method~\cite{muravyev2025prism}. We build the map $M=(V, E)$, where $V = (v_1, \dots, v_N)$ are the locations with their ground truth poses $p_1, \dots, p_N$ respectively, and $E = \{(v_i \in V, v_j \in V):\ ||p_i - p_j|| < th\}$ are the edges between locations (we link a pair of locations only if those observation points are located not further than a certain threshold from each other). As the relative poses for graph edges, we use the difference between ground truth poses of the locations. For each location, we store the 2D grid extracted from its point cloud by 2D projection, and the descriptor for place recognition. So, each location consumes only several kilobytes in disk storage, which lets us store even city-scale topological maps onboard. An example of a topological map is shown in Fig.~\ref{fig:topomap}.

\subsection{Localization}

\begin{figure*}[ht]
    \vspace*{0.15in}
    \centering
    \includegraphics[width=0.9\textwidth]{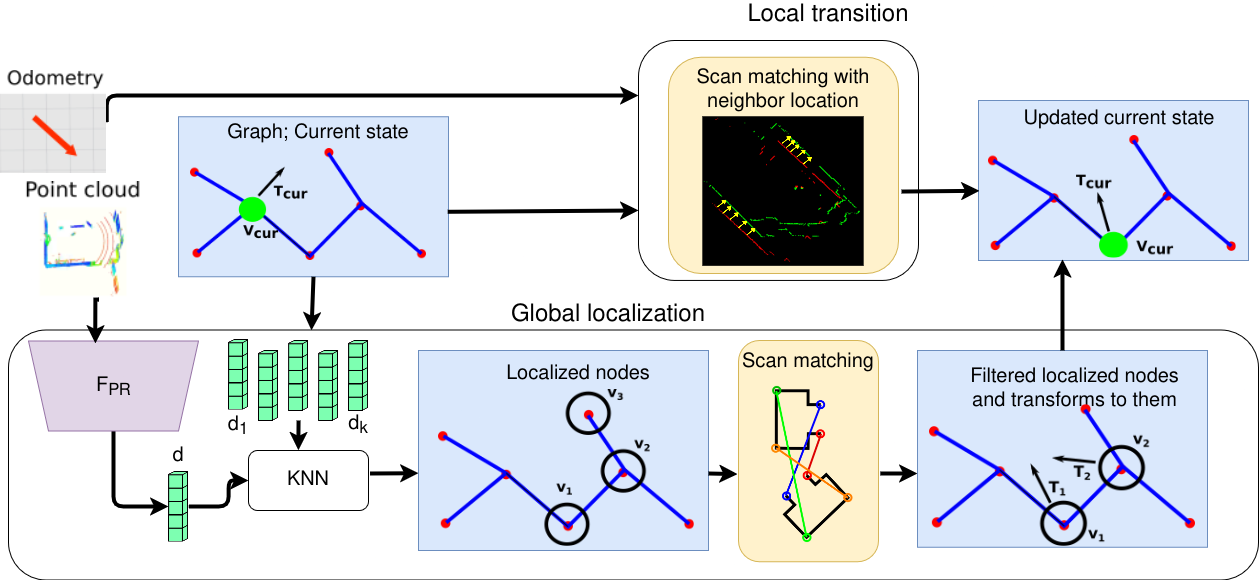}
    \caption{A scheme of the proposed global localization pipeline. It includes $F_{PR}$ place encoder (we use MinkLoc3D) and the proposed scan matching module. The output of the pipeline is the current location and the robot's relative pose in this location.}
    \label{fig:localization_pipeline}
\end{figure*}

\subsubsection{Overall pipeline}

The proposed localization pipeline maintains current state in the graph of locations. At each step $t$, this state is represented as the location $v_{cur}^t$ where the robot is currently located, and the robot's relative position $T_{cur}^t$ in this location. The pipeline is similar to the graph maintaining pipeline of the original PRISM-TopoMap method~\cite{muravyev2025prism}, however, it has no part of a new node addition and checks the localization results for proximity to the current state:

\begin{itemize}
    \item First, we check whether the robot is still inside $v_{cur}^{t-1}$ and its current scan overlaps with $v_{cur}^{t-1}$ by a sufficient percent. If the check passes, we apply the odometry measurement to $T_{cur}$: $v_{cur}^t = v_{cur}^{t-1}; T_{cur}^t = T_{cur}^{t-1} \cdot o_t.$

    \item If we moved outside $v_{cur}^{t-1}$, or our overlapping percent is low, we first try to change $v_{cur}$ to one of its neighbors in the graph (i.e., move along an edge). For this purpose, we use the approximate relative position $((T_{cur}^{t-1})^{-1} (p_{v_{cur}})^{-1} p_{v_{next}})^{-1}$ as an initial guess for scan matching. If we matched successfully to one the neighbors $v_{next}$, then we change $v_{cur}$ to it: $v_{cur}^t = v_{next}; T_{cur}^t = T_{next}$. To prevent false matches, we use a "jumping threshold" -- the maximum distance between an initial guess and the found transform when the matching is considered successful.
    \item If we could not move along an edge, we try to change $v_{cur}$ using the localization results. If there is found the location $v_{loc}$ with sufficient overlap, we change $v_{cur}$ to it: $v_{cur}^t = v_{loc}; T_{cur}^t = T_{loc}$.
    \item After all, if we could not change $v_{cur}$ by edges and by the localization results, we change its value to the nearest neighbor in the graph without transform alignment (i.e., try to move along an edge without matching): $v_{cur}^t = v_{next}; T_{cur}^t = ((T_{cur}^{t-1})^{-1} (p_{v_{cur}})^{-1} p_{v_{next}})^{-1}$. We do this change if $||((T_{cur}^{t-1})^{-1} (p_{v_{cur}})^{-1} p_{v_{next}})^{-1}||<5$ to stay inside the next location. Otherwise, we do not change $v_{cur}$, apply step (1), and consider that the localization is lost.
\end{itemize}

The estimated global position in each time step is simply the application $T_{cur}^t$ to $v_{cur}^t$'s pose:

$$\widehat{p_{t}^{next}} = p_{v_{cur}^t}^{first} T_{cur}^t.$$

For global localization in the topological map, we use a twofold pipeline which first searches the proper location in the map using place recognition, and then refines the place recognition results and provides the precise pose estimation using the proposed scan matching algorithm. The scheme of the pipeline is depicted in Fig.~\ref{fig:localization_pipeline}.

\subsubsection{Place Recognition}


For place recognition we use MinkLoc3D~\cite{komorowski2021MinkLoc3DPointCloud}, selected after training and comparing several models on NCLT --- a large $360\deg$ LiDAR dataset close to our domain and to ITLP-Campus \cite{melekhin2024itlp} --- where it achieved the best performance (Table~\ref{tab:pr_comparison});
however, due to noisy GT trajectories, NCLT is unsuitable for strict end-to-end evaluation of the full topological-localization pipeline.
For each robot scan, we retrieve the top-5 nearest locations in the topological map by Euclidean distance between MinkLoc3D descriptors and pass these candidates to subsequent pipeline stages.

\begin{table}[ht]
    \caption{PR performance on the NCLT \cite{carlevaris2016university} dataset}
    \setlength\tabcolsep{10pt}
    \label{tab:pr_comparison}
    \centering
    \begin{tabular}{l|ccc}
        \hline
        \multirow{2}{*}{Method} & \multicolumn{2}{|c}{NCLT} & Inference \\
         & R@1 & R@5 & time, ms \\
        \hline
        MinkLoc3D \cite{komorowski2021MinkLoc3DPointCloud} & \textbf{90.5} & \textbf{97.1} & \\
        ASVT-Net \cite{fan2022SVTNetSuperLightWeight} & \underline{86.9} & \underline{96.0} & \\
        CSVT-Net \cite{fan2022SVTNetSuperLightWeight} & 85.9 & 95.2 & \\
        SVT-Net \cite{fan2022SVTNetSuperLightWeight} & 81.9 & 94.6 & \\
        \hline
    \end{tabular}
\end{table}

\subsubsection{Scan Matching}

Learning-based place recognition methods sometimes output false positive matches due to visual similarity of different parts of the environment. Therefore, relying solely on a place recognition network can cause our topological mapping method to link far, non-adjacent locations. To prevent this, we filter place recognition results matching the scan from the robot with the scans of the found locations, thereby finding relative transforms between these scans.

To estimate the transform between the reference scan $R$ from the robot, and candidate scan $C$ from the map, we combine three techniques. First, we match the scans using the features $R_f$ and $C_f$ extracted by ORB detector~\cite{rublee2011orb} (for global localization) and by Harris corner feature detector~\cite{harris1988combined} for matching along an edge. We find the correspondences between features using FLANN matcher~\cite{muja2009flann}, as in the PRISM-TopoMap method.


Based on the identified correspondences, we apply a modified RANSAC algorithm to estimate rotation and translation. First, the features from the candidate map are transformed into the reference frame using odometry. For each correspondence, we assign a prior probability $p_i$ proportional to $\exp\!\left(-d^2 / \kappa^2\right)$ , where $d$ is the distance between the transformed feature from the candidate map and its corresponding feature in the reference frame and $\kappa = 1 \ m$. These prior probabilities $p_i$ are then used in the sampling step to improve robustness and reduce the number of required iterations.



\begin{figure}[ht]
    \centering
    \includegraphics[width=0.48\textwidth]{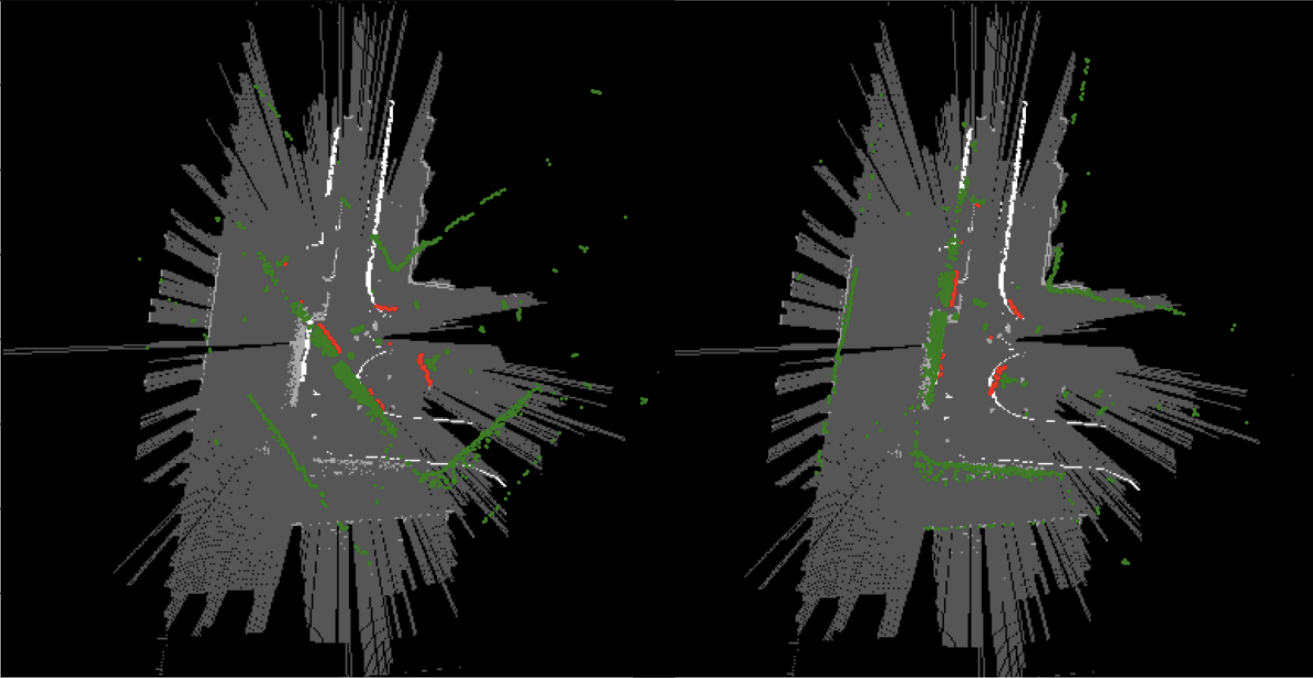}
    \caption{Scan matching comparison: our previous algorithm (left) \cite{muravyev2025prism} and the proposed algorithm (right). The candidate scan is shown in gray and white, the reference scan is shown in green (wall points) and red (curb points)}
    \label{fig:compare_with_old}
\end{figure}

An example of the comparison between the proposed scan matching and the PRISM-TopoMap's one is shown in Fig. \ref{fig:compare_with_old}.

\begin{figure*}[h!]
    \centering
    \includegraphics[width=1\textwidth]{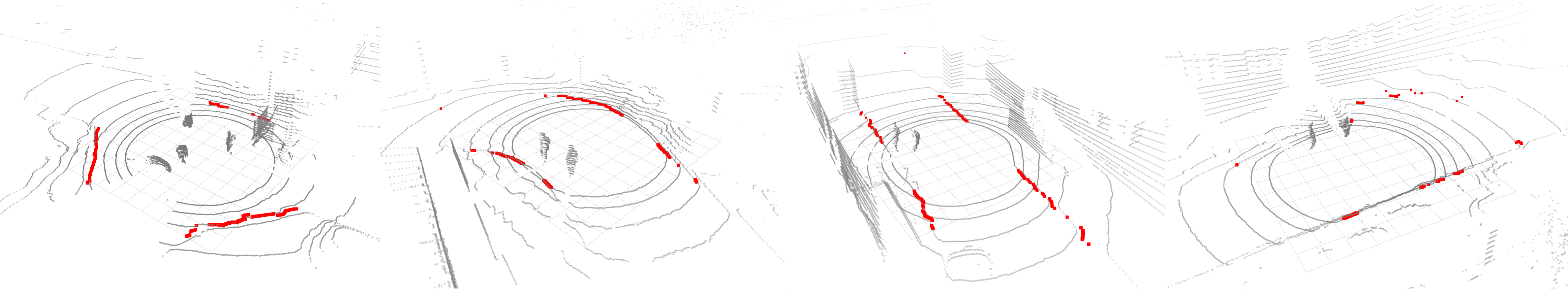}
    \caption{Curb detection output. Points belonging to curbs are shown in red. }
    \label{fig:curbs_example}
\end{figure*}

\subsubsection{Curb Detection}

The key idea of the curb detection algorithm is to accurately segment the ground (drivable region) through plane fitting, from which curbs are derived as edges of this region. The process begins by applying a sliding window across the azimuth to compute the standard deviation of the radial distance (range) for each point. Points exhibiting a low standard deviation are retained, as they are likely to belong to flat surfaces such as the ground. Subsequently, points sampled from the previous step are weighted with a higher probability near the center line to minimize the influence of pavements on the sides. Plane fitting is then performed on these filtered points using the RANSAC algorithm. The original point cloud is rotated to align the vertical axis with the normal of the fitted plane.
The ground region is defined as points lying below a threshold of 0.1 m and is represented as a binary 2D array (mask) in the range view. Edges within the ground region mask are detected while ignoring points near regions with heights exceeding 0.3 m to eliminate noise from dynamic objects such as pedestrians. Finally, the resulting indices of edges in the range view are used to reconstruct the 3D coordinates of the detected curbs in the robot`s frame. Examples of curb detection algorithm output is shown in Fig. \ref{fig:curbs_example}.

\begin{algorithm}
\caption{Curb Detection}\label{alg:curbs_surface_fit}
\KwData{Point cloud: $P$}
\KwResult{Points of curbs: $C$}
\tcp{{Ground plane segmentation}}
$P_{f} \gets rangeSTDWindowFilter(P, \sigma_{threshold}^2)$\;
$P_{s} \gets sample(P_f, \sigma_{y}^2)$\;
$T_{pb} \gets ransacFitPlane(P_{s}, d_{threshold})$\;
$P_{t} \gets transform(P, T_{pb})$\;
\BlankLine
\tcp{{Edges detection}}
$P_u \gets upsample(P_t, f=3)$\;
\tcp{Mask in range view}
$mask_1 \gets P_u.z < \text{curb\_height}$ \;
$mask_2 \gets abs(P_u.z) > 3 \cdot \text{curb\_height}$\;
$mask_2 \gets dilate(mask_2, ones(w=50, h=7))$\;
$mask \gets \neg mask_2 \wedge mask_1$\;
$C \gets edges(P_u, mask)$
\end{algorithm}

\section{EXPERIMENTS}
\label{sec:experiments}

\subsection{Setup and Datasets}

We evaluated the proposed method and the competitors on three large-scale outdoor datasets. The first of them is the outdoor part of the ITLP-Campus dataset~\cite{melekhin2024itlp}. This dataset contains several runs of a mobile differential-drive Clearpath Husky robot on a university campus. These runs were recorded in different seasons and daytimes, and all of them cover the whole campus. The sensor data include scans from a 16-ray Velodyne LiDAR, images from two cameras, and wheel encoders' data. 

The second dataset includes the first 6 sequences of Oxford RobotCar Radar~\cite{barnes2020oxford}, which were collected from an autonomous car in an urban environment. It contains data from two 32-ray LiDARs, IMU, and two cameras. The summary of used datasets is shown in Table~\ref{tab:datasets}.

\begin{table}[ht!]
    \caption{Datasets used in the experiments}
    \setlength\tabcolsep{2pt}
    \label{tab:datasets}
    \centering
    \begin{tabular}{l|cccc}
        \hline
        Dataset & N of seqs & Environment & \begin{tabular}{c}Avg. track\\length, km\end{tabular} & IMU\\
        \hline 
        ITLP-Campus~\cite{melekhin2024itlp} & 4 & Campus & 3.3 & No\\
        RobotCar~\cite{barnes2020oxford} & 6 & Urban & 7.8 & Yes\\
        \hline
    \end{tabular}
\end{table}


For both datasets, we used one track for map creation, and the rest of the tracks were used for evaluation of the localization within the built map. For ITLP-Campus dataset, we created the map by track 03, and ran localization on track 00, 01, and 02. The main challenge of evaluation on this dataset is the seasonal difference between tracks (tracks 00 and 01 were recorded in winter and have a large amount of snow piles on lidar scans, and tracks 02 and 03 were recorded in spring). For Oxford RobotCar, we created the map by track 2019-01-11-14-02-26, and evaluated the localization methods on five other tracks recorded in January 2019. The main challenge of this dataset is the presence of a large amount of dynamic objects (like cars and pedestrians), which may obstruct lidar scan matching.


\subsection{Baselines}
We used 2 methods to compare with the proposed method.

The first one is a classical point cloud-based approach, where the algorithm operates with a point cloud itself. For localization in such methods, an accurate and dense point cloud should be given as an input.
To create a 3D point cloud map, DLO~\cite{chen2022direct} method was used. It consists of two components, odometry estimation and 3D mapping. It cannot work in localization-only mode (only in SLAM mode), so, for localization in the built map, we used HDL\_Localization method~\cite{koide_Portable_2019}.

DLO is a lightweight 3D LiDAR-inertial SLAM method. For odometry estimation, it uses NanoGICP (an optimized version of Generalized ICP~\cite{koide2021voxelized} algorithm). Global 3D map is built as a set of 3D submaps, each submap is created from a single voxelized LiDAR scan. Such solution allows to create compact 3D maps for large environments. To build precise map relying on ground truth positions with DLO, we used these positions instead of DLO odometry estimations.


Another method was BEVPlace++ \cite{luo_BEVPlace_2024}. It introduces a fast, robust, and lightweight LiDAR global localization method designed for unmanned ground vehicles. The approach begins by projecting sparse LiDAR point clouds into bird’s-eye view (BEV) images, which serve as a compact and informative representation. A specially designed Rotation Equivariant and Invariant Network (REIN) then extracts rotation-equivariant local features and rotation-invariant global descriptors from these BEV images. This enables a two-stage localization process where coarse place recognition is performed via global descriptors and refined through local feature matching with RANSAC for accurate 3-DoF pose estimation. The method’s efficiency, combined with its robustness to view and sensor variations without requiring dense pose supervision, makes it an appealing baseline for real-time global localization tasks.

\begin{figure}[ht!]
    \begin{subfigure}[t]{0.21\textwidth}
\centering\includegraphics[height=4cm,width=1\textwidth,keepaspectratio]{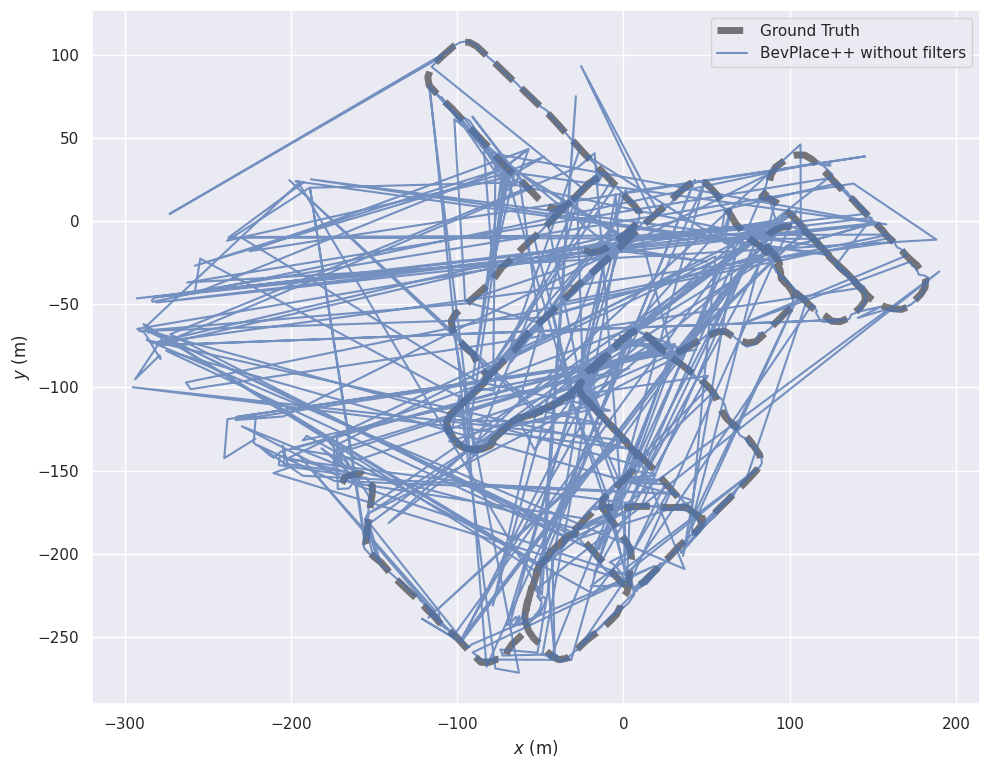}
        \caption{Trajectory on $XY$ plane}
    \end{subfigure}
    \begin{subfigure}[t]{0.27\textwidth}
\centering\includegraphics[height=4cm,width=1\textwidth,keepaspectratio]{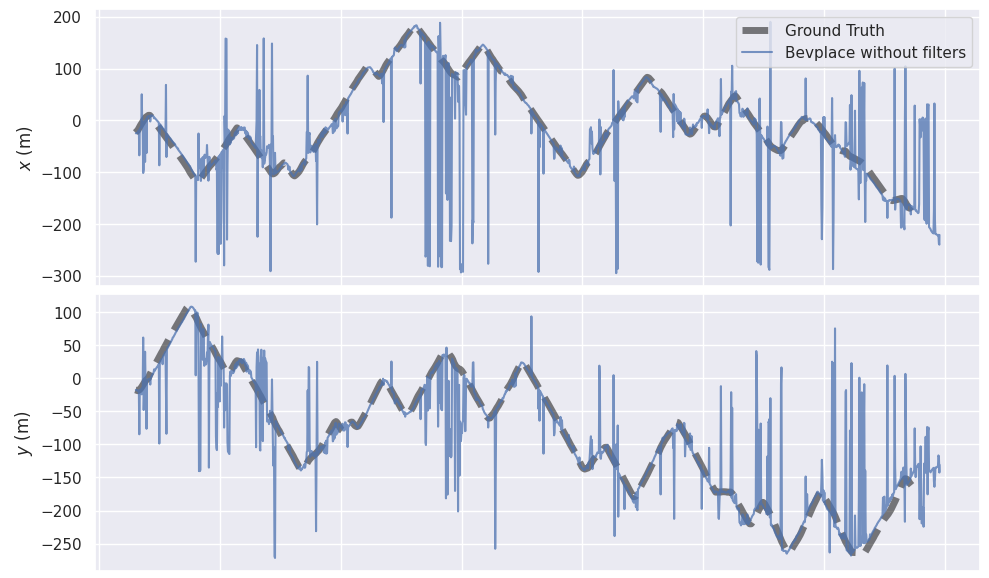}
        \caption{Coordinates respect to time}
    \end{subfigure}

\caption{BEVPlace++ method without applied filters, track 02}
\label{fig:bevplace_wo_filters}
\end{figure}

A significant limitation of the BEVPlace++ algorithm stems from its dependence on an approximate initial pose estimate. Specifically, the method requires prior knowledge of the vehicle’s rough global position at each timestamp to retrieve the optimal BEV reference image from the precomputed database. While this approach is effective for refining localization in systems equipped with supplementary sensors (e.g., wheel odometry), it inherently restricts the method’s applicability to global localization tasks using exclusively LiDAR-derived data. As illustrated in Figure~\ref{fig:bevplace_wo_filters}, unfiltered BEVPlace++ exhibits an average Absolute Trajectory Error (ATE) Root Mean Square Error (RMSE) of approximately 73 meters across test trajectories, underscoring the necessity for robust filtering mechanisms.

As a possible improvement, we consider a lightweight temporal-coherence filtering strategy that leverages the “last known database position index”—assuming sequential database indices correspond to nearby poses. This option reduces reliance on external localization modalities (e.g., odometry) by exploiting temporal continuity and the approximate inter-frame distance between consecutive database entries.

\begin{table*}[ht!]
    \caption{Localization quality values}
    \setlength\tabcolsep{4pt}
    \label{tab:results}
    \centering
    \begin{tabular}{l|ccc|ccc}
        \hline
        \multirow{2}{*}{Method} & 
        \multicolumn{3}{|c}{ITLP-Campus} &
        \multicolumn{3}{|c}{RobotCar} \\
        \cline{2-7}
         & Mean ATE & Median ATE & $SR_{loc}$ &
         Mean ATE & Median ATE & $SR_{loc}$ \\
        \hline
        GLIM + HDL loc. &
        \multicolumn{2}{c}{failed} & 0.04 &
        4.3* & 1.9* & 0.72 \\
        BEVPlace++ &
        14.4 & 11.0 & 0.84 &
        14.9 & 1.6 & 0.94 \\
        \hline
        [ours] PRISM-Loc &
        0.5 & 0.3 & 0.99 &
        5.9 & 1.8 & 0.90 \\
        \hline
    \multicolumn{7}{l}{\footnotesize* Localization failed on tracks 2019-01-15-13-06-37 and 2019-01-18-15-20-12;}\\
    \multicolumn{7}{l}{\quad ATE reported only up to failure}\\
    \end{tabular}
\end{table*}

\begin{table*}[ht!]
    \caption{Localization performance values}
    \setlength\tabcolsep{4pt}
    \label{tab:results_performance}
    \centering
    \begin{tabular}{l|ccc|ccc|ccc}
        \hline
        \multirow{2}{*}{Method} &
        \multicolumn{3}{|c}{ITLP-Campus} &
        \multicolumn{3}{|c}{RobotCar} \\
        \cline{2-7}
         & RAM, GB & Map size, GB & Runtime, s &
         RAM, GB & Map size, GB & Runtime, s \\
        \hline
        DLO + HDL loc. &
        1.2 & 0.05 & 1.0 &
        4.2 & 0.12 & 1.0\\
        BEVPlace++ &
        1.5 & 0.8 & 0.4 &
        2.6 & 10.6 & 0.4 \\
        \hline
        [ours] PRISM-Loc &
        0.2 & 0.02 & 0.3 &
        0.7 & 0.02 & 0.3\\
        \hline
    \end{tabular}
\end{table*}

Concretely, the filter constrains the search space to indices adjacent to the last confirmed position: FAISS-retrieved candidates are re-scored by their proximity (in index space) to the prior validated index. This increases robustness to isolated outliers, though it does not handle loop closures and may still lead to track loss.

Two baseline frameworks were evaluated against the proposed method: (1) HDL localization based on a DLO-created map, and (2) BEVPlace++ (optionally augmented with the temporal-coherence filter) developed for rough real-time pose estimation.
 
\subsection{Results}



We recorded the positions estimated by each of the methods and measured the metrics \ref{eq:ate_mean}\ -\ \ref{eq:sr_loc} for them, as well as performance values, such as the map size in the disk storage, memory usage during the localization, and the average time spent to one global localization iteration. For computation of localization success rate, we use the threshold of 10 m.

\begin{figure}[h]
\centering\includegraphics[height=8.5cm,width=1\textwidth,keepaspectratio]{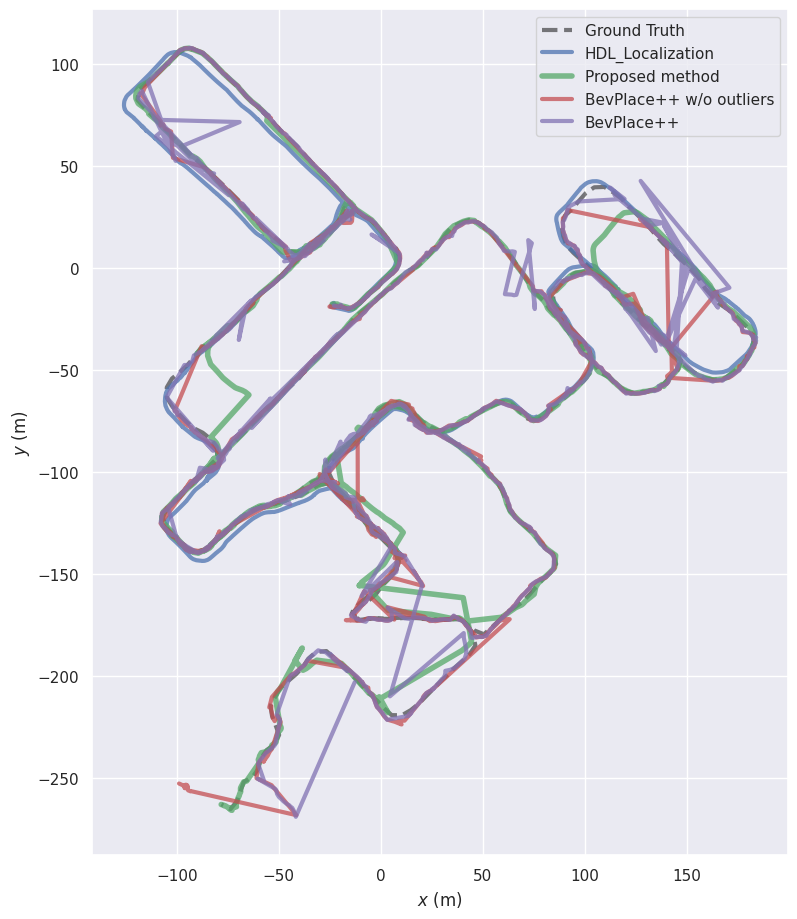}
        \caption{Trajectory on $XY$ plane}
    

\label{fig:track00}
\end{figure}

The results are shown in Table \ref{tab:results}. The comparison of the estimated trajectories with all the methods and the boxplots for trajectory errors are shown in Fig.~\ref{fig:track00}. At each track of ITLP-Campus, HDL method failed to localize in the 3D map after traveling about 200 m. The sparse voxelized point cloud map for the 3 km route weighed only about 50 MB; however, each localization procedure in it took about 1 s, with complete localization loss in the beginning of the track. On dense urban environment of Oxford RobotCar dataset, HDL failed on two tracks of five, reaching 72\% success rate, and 4 m mean ATE on succeeded tracks.

Because of significant changes between the tracks, the pure BEVPlace++ method without filtering exhibited about 16\% of place recognition errors, which led to a mean ATE of 20 m (see Fig. \ref{fig:bevplace_wo_filters}). The suggested filtering approach eliminated most of the place recognition errors, however, the percent of localization fails remained high. The map for BEVPlace++ took 15 GB on disk storage and in the RAM because of storing large local descriptors for each location. Also, the localization pipeline took in average 1.0 sec (about 700 ms for place recognition, and 300 ms for pose estimation using a local descriptor).

The proposed PRISM-Loc method made localization losses and errors in only 1\% of cases on ITLP and in 10\% of cases on Oxford RobotCar thanks to our combined approach to localization. The mean ATE on ITLP-Campus the value of 0.5 m, significantly outperforming all the competitors. With such localization quality, our approach remained fast and lightweight -- the maps for both ITLP and Oxford RobotCar consumed only 20 MB (the largest part of 4 GB RAM consumption is the reserving allocation for PyTorch models, this memory is not utilized, so memory usage could be optimized in the future). Also, the global localization time took only 0.3 s (0.1 s for place recognition, and 0.04 s for scan matching with each of five found candidates).

\FloatBarrier

\section{CONCLUSION AND FUTURE WORK}
\label{sec:conclusion}

We propose PRISM-Loc -- a lightweight and robust method for localization in large outdoor environments using topological maps. The proposed method combines tracking through rough odometry estimations, a novel robust scan-matching technique integrated with the original curb detection algorithm, and place recognition for global location search. Experiments conducted on two challenging large-scale datasets with route lengths up to 8 km demonstrate that the proposed method achieves robust localization, with a mean trajectory error of 0.5 meters, despite significant object-wise and seasonal differences between the run and the map. Furthermore, PRISM-Loc maintains low memory usage and computational time throughout the entire route. Moving forward, we plan to enhance our approach by incorporating local pose filtering to improve robustness and to conduct extensive research on scan-matching techniques.




\bibliography{IEEEexample}
\bibliographystyle{myIEEEtran}

\end{document}